\documentclass[sigconf]{aamas}  

\usepackage{booktabs}
\usepackage{tikz}
\pgfdeclarelayer{background}
\pgfdeclarelayer{foreground}
\pgfsetlayers{background,main,foreground}
\usepackage{pgfplots}
\definecolor{Gray}{gray}{0.9}
\usepackage{color, colortbl}
\usetikzlibrary{patterns}

\usepackage{flushend}
\setcopyright{ifaamas}  
\acmDOI{doi}  
\acmISBN{}  
\acmConference[AAMAS'20]{Proc.\@ of the 19th International Conference on Autonomous Agents and Multiagent Systems (AAMAS 2020), B.~An, N.~Yorke-Smith, A.~El~Fallah~Seghrouchni, G.~Sukthankar (eds.)}{May 2020}{Auckland, New Zealand}  
\acmYear{2020}  
\copyrightyear{2020}  
\acmPrice{}  

\begin{document}

\title{Optimising Game Tactics for Football}

\author{Ryan Beal\textsuperscript{\rm 1}, Georgios Chalkiadakis\textsuperscript{\rm 2}, Timothy J. Norman\textsuperscript{\rm 1} and Sarvapali D. Ramchurn\textsuperscript{\rm 1} \\
\textsuperscript{\rm 1} University of Southampton, United Kingdom, \{ryan.beal,t.j.norman, sdr1 \}@soton.ac.uk \\
\textsuperscript{\rm 2} Technical University of Crete, Greece, gehalk@intelligence.tuc.gr}

\renewcommand{\shortauthors}{R. Beal et al.}

\begin{abstract}  

In this paper we present a novel approach to optimise tactical and strategic decision making in football (soccer). We model the game of football as a multi-stage game which is made up from a Bayesian game to model the pre-match decisions and a stochastic game to model the in-match state transitions and decisions. Using this formulation, we propose a method to predict the probability of game outcomes and the payoffs of team actions. Building upon this, we develop algorithms to optimise team formation and in-game tactics with different objectives. Empirical evaluation of our approach on real-world datasets from 760 matches shows that by using optimised tactics from our Bayesian and stochastic games, we can increase a team chances of winning by up to 16.1\% and 3.4\% respectively. 

\end{abstract}


\maketitle


\section{Introduction}\label{sec:intro}
Many real-world settings can be modelled as games involving a pair of teams.   In these types of games, each team optimises its tactical decisions to maximise its chances of winning. Examples include politics where teams of politicians aim to win an election as a party \cite{snidal1985game} and  defence and security, where teams of agents schedule their rota to protect a facility against attackers (e.g., Stackelberg Games \cite{paruchuri2008playing} and Port Security \cite{shieh2012protect}).  In this paper we focus on a model for games in team sports, specifically on games of Association Football (soccer).\footnote{Referred to as just ``football'' throughout this paper.} The popularity of football has grown significantly over the last 20 years and is now a huge industry in the world providing 1 million jobs in the UK alone and entertainment to over a billion people worldwide. According to a recent report\footnote{https://www2.deloitte.com/uk/en/pages/press-releases/articles/annual-review-of-football-finance-2018.html.} in 2018,  the estimated size of the European sports industry is €25.5 billion. 

 In football, there are two teams (made up of 11 individual players) who work together to score and prevent goals. Winning a game involves many tactical decisions, including but not limited to, assigning positions to players, composing a team, and reacting to in-game events. Such decisions have to be made against significant degrees of uncertainty, and often in very dynamic settings.  Prior multi-agents research for football has focused more on the contribution of individual agents within a team \cite{decroos2019actions, beal2019teams}. However, to date, there is no formal model for the tactical decisions and actions to improve a team's probability of winning. There are a number of tactical decisions that are made both pre-match and during the match. These decisions are often just made through subjective opinions and ``gut feelings'', and are a popular subject of discussion in football media.\footnote{https://www.bbc.co.uk/sport/football/50334352.} 

Against this background, we propose a formal model for the game of football and the tactical decisions that are made in the game. We model the game as a 2-step game that is made up of a Bayesian game to represent the pre-match tactical decisions that are made due to the incomplete information regarding the tactical choices of the opposition. We then use a stochastic game to model the state-transitions in football and the tactical decisions that can be made during the match. It is important to note how the decisions in both the Bayesian and stochastic game feed back into one another as the pre-match decisions impact the starting strategies during the game, and the in-game decisions allow us to learn what tactics work well against certain teams. This formal model allows us to learn the payoffs of given decisions and actions so that we can optimise the decisions that are made by a team. We validate and test our model and algorithms to data from real-world matches. We show that our pre-match and in-match tactical optimisation can boost a team's chances of obtaining a positive result from a game (win or draw). We also show that we can use machine learning effectively to learn pre-match probabilities based on tactical decisions and accurately predict state changes. Thus, this paper advances the state of the art in the following ways:

\begin{enumerate}
    \item We propose a novel mathematical model for the game of football and the tactical decision-making process.
    \item Using real-world data from 760 real-world football games from the past two seasons of the English Premier League (EPL), we can learn the payoffs for different team actions and learn state transitions. In particular, we show that we can predict game-state transitions with an accuracy of up to 90\%.  We also show we can accurately predict opposition tactical decisions.
       \item By learning action payoffs, we can optimise pre- and in-match tactical decisions to improve the probability of winning a game.
\end{enumerate}

When taken together, our results establish benchmarks for a computational model of football and data-driven tactical decision making in team sports. Furthermore, our work opens up a new area of research into the use of these techniques to better understand
how humans make decisions in sport.

The rest of this paper is organised as follows. In Section \ref{sec:background} we review the literature on tactics in football. Section \ref{sec:model} formally defines our model and tactical optimisation. In Sections \ref{sec:solve-bg} and \ref{sec:in-game} we discuss the methods we use to optimise decisions for our Bayesian game and stochastic game respectively. In Section \ref{sec:experiments} we validate and test our models using real-world datasets. Finally, in Section \ref{sec:discussion} we discuss out results and in Section \ref{sec:conclusions} we conclude.

\section{Background}\label{sec:background}

In this section, we review related literature showing applications of game theory to real-world problems and give an overview of why football tactics are important and what they involve.

\subsection{Related Work}\label{subsec:lit}
We discuss examples of work that models real-world strategies as well as work relating to decision making in football.

\subsubsection{Modelling Real-world Strategic Interactions:} Work in \cite{silver2016mastering} models the game of Go and makes strategic decisions in the game using deep neural networks and tree search. Their program (AlphaGo) achieved a 99.8\% winning rate against other Go programs, and defeated the human European Go champion by 5 games to 0. In \cite{synnaeve2011bayesian} Bayesian modelling is used to predict the opening strategy of opposition players in a real-time strategy game. They test their model on a video game called StarCraft.\footnote{StarCraft and its expansion StarCraft: Brood War are trademarks of Blizzard Entertainment} This approach is similar to how we aim to predict the strategy of the opposition in our Bayesian game. However, in our approach, we look to take this one step further and optimise our strategy around the potential strategies and opposition could take and then feed this into a stochastic game \cite{shapley1953stochastic}. Other examples of work for opponent modelling are shown in \cite{billings1998opponent,carmel1995opponent,hsieh2008building}. 

There are also examples of strategic models in security. Work in \cite{paruchuri2008playing} focuses on Bayesian Stackelberg games, in which the player is uncertain about the type of the adversary it may face. This paper presents efficient algorithms using MIP-formulations for the games and test their approaches on several different patrolling games. Following on from this there are applications of game theoretic approaches for security shown in \cite{shieh2012protect}. In this work, the authors present a game-theoretic system deployed by the United States Coast Guard in the port of Boston for scheduling their patrols. This paper shows an example of a model for a real maritime patrolling problem and test this using real-world data (using mock attacks).

In \cite{chen2013synthesis} a model is developed for strategies in two-player stochastic games with multiple objectives explored for autonomous vehicles and stopping games. This shows one of the first applications of multi-objective stochastic two-player games. Another example of a model for stochastic games is shown in \cite{kardecs2011discounted} which shows the use of discounted robust stochastic games in a single server queuing control. Finally, work in \cite{avsar2002inventory} models the problem of inventory control at a retailer formulating the problem as a two‐person nonzero‐sum stochastic game. 

The work in this paper differs from previous work as we use real-world data for a real-world problem creating a novel model that can feed optimised strategies from a pre-match Bayesian game into an in-match stochastic game; to the best of our knowledge, this is the first time such an intertwining is proposed in the literature.

\subsubsection{Decision-Making in Sport:} There has also been work in the sports domain focusing on tactics and looking at game-states in football. Firstly, work in \cite{jordan2009optimizing} explores different risk strategies for play-calling in American Football (NFL). Although game theory has not been applied to football tactics in prior work, some examples of key work to understand the game better are shown in \cite{le2017datadriven}. There, deep imitation learning has been used to ``ghost'' teams so that a team can compare the movements of its players to the league average or the top teams in the league. Also, \cite{fernandez2019decomposing} provides a model to assess the expected ball possession that each team should have in a game of football. These papers help to identify where teams can make changes to their styles of play to improve their tactics. Another application of learning techniques in football is shown for fantasy football games in \cite{matthews2012competing}. 

To give more background around these papers and the problem we are looking to solve, in the next subsection we give a background to football tactics and their importance to the game.

\subsection{Football Tactics}\label{subsec:ft}

\begin{figure}[h!]
        \centering
        \includegraphics[scale=0.4]{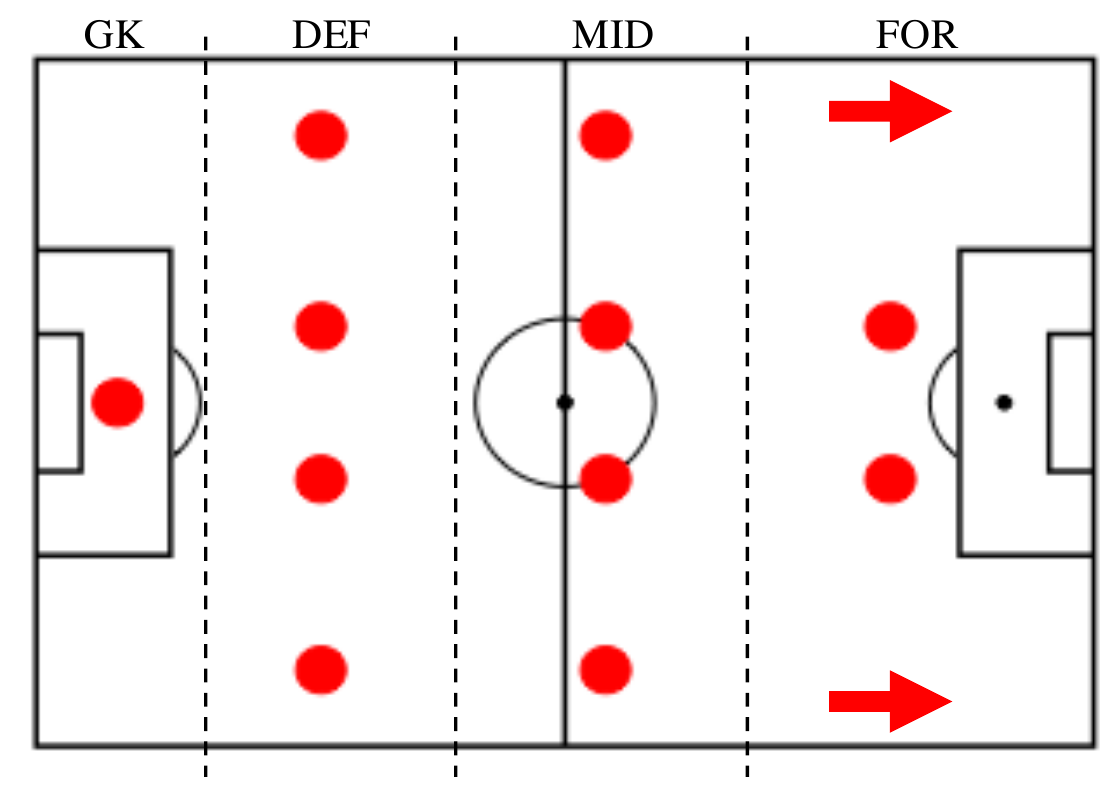}
        \caption{\small Example Team Formation (4-4-2).}
        \label{fig:form}
\end{figure}

 The foundations of sports and football tactics are discussed in \cite{grehaigne1999foundations,bate1988football} and applications of AI is discussed in \cite{beal2019artificial}. There are multiple tactical decisions that a coach or manager must make before and during a game of football. These can have a significant impact on the overall result of the game and can help boost the chance of a team winning, even if a team does not have the best players. It is worth noting that in a standard league (such as the EPL or La Liga) a win is worth 3 points, a draw 1 point and a loss no points. Therefore, some teams aim to pick more reserved tactics to increase their chances of drawing a game which they are unlikely to win. Managers and coaches prepare for their upcoming matches tactically to the finest details, usually by using subjective opinions of their own and opposition team/players. Some of the key pre-game decisions that are made by both teams include: 

\begin{itemize}
    \item \textbf{Team Style:} A teams playing style is a subjective concept that relates to the teams overall use of different playing methods. There are many different styles that a team can use but these can be analysed using game statistics and similar styles can be identified. Some examples of these are shown in Table \ref{tab:style}.\footnote{More styles explained here: https://www.90min.com/posts/2990047-6-very-different-styles-of-football-from-across-the-world.}
    \item \textbf{Team Formation:} The formation is how the players are organised on the pitch. There is always 1 goalkeeper and 10 outfield players who are split into defenders (DEF), midfielders (MID) and forwards (FOR). An example of a formation is 4-4-2, this represents 4 defenders, 4 midfielders and 2 forwards. Figure \ref{fig:form} shows how this is set up on the pitch (attacking in the direction of the arrows). 
    \item \textbf{Selected Players:} The selected players are the 11 players that are selected to play in the given starting formation or selected to be on the substitute bench (between 5-7 players). Some players may perform better in different styles/formation or against certain teams.
\end{itemize}

\begin{table}[h!]
\centering
\begin{tabular}{|c|c|}
\hline
\rowcolor{Gray}
\textbf{Style} & \textbf{Description} \\ \hline
Tika-Taka &   Attacking play with short passes.  \\ \hline
Route One & Defensive play with long passes.   \\ \hline
High Pressure & Attack by pressuring the opposition.  \\ \hline
Park The Bus & A contained defensive style. \\ \hline
\end{tabular}
\caption{Example Playing Styles.}
\label{tab:style}
\end{table}

In terms of the in-game decisions that are made, one change that can be made is with a substitution (other examples include tweaks to the style and formation). Therefore, we can model the in-game decisions as a stochastic game and look to make optimised substitutions that increase the probability of scoring a goal. This can help teams to improve their chances of winning games and learn from historic datasets.

Due to the number of decisions that can be made by teams in football both before and during the match, there are many uncertainties both in what the opponent may do and on how the decisions made may affect the overall outcome of the game. In this paper, we aim to address some of these uncertainties so we can optimise tactical decision-making in football. In the next section, we define the model we use to achieve this.

\section{A Formal Model for the Game of Football}\label{sec:model}
We model the tactical decisions that are made in football into two parts. First, we model the pre-match tactical decision making process as a Bayesian game, taking into account the fact that each team has incomplete information regarding the opposition's tactical decisions before the game begins. Second, we model the in-match decisions as a stochastic game due to changing states of a game of football as the game progresses (see Section \ref{subsec:in-game} for more details on the states of the game). We use these  two as modelling tools and frameworks to conduct learning in, rather than aiming to solve for equilibria. This is because ours is a real-world dynamic setting with multiple sources of uncertainty; and one in which the identity of the opponents faced by a player changes for every instance of the game (moreover, the same opponent is met only twice per season). Thus, solving for some kind of equilibrium of the overall setting is clearly impractical. By contrast, opponent types and probabilistic transitions among states that represent scorelines given formation and style are natural in this problem. As such, our Bayesian and stochastic game framework provides a natural model to facilitate learning in this domain.


\subsection{Pre-Match Bayesian Game}\label{subsec:pre-game}

As discussed in Section \ref{subsec:ft}, there are many unknown factors about the opposition and many decisions that must be made for a team to maximise their chances of having a positive outcome from a game. Therefore, we use a Bayesian game to model strategic decisions and make optimised decisions. In our Bayesian game we define the two teams as $T = \{T_\alpha$, $T_\beta\}$ where $T_\alpha$ is the team whose actions we are optimising and $T_\beta$ is the opposing team. Each of these teams has a corresponding action set $A_\alpha$ and $A_\beta$. These are sets of one-shot actions selected by both teams, involving tactical choices before the match (i.e. selecting the team formation is a single decision and selecting the starting 11 players are 11 separate decisions for each position). Each individual action/decision is $a \in A$.

There is a set of possible types $\Theta$ that a team can select and each team's corresponding type is defined as $\theta_\alpha$ and $\theta_\beta$ where $\theta \in \Theta$. These types correspond to the style of football in which an opposition is likely to use (e.g., tika-taka, route one and high pressure). Some other examples of how we can use team types in football would include: the strength of the team (in terms of league position) and the difference between home and away teams. We then have a probability function which represents a teams' prior beliefs about the choices of their opposition regarding their style type and the actions that they may take (in this case it is the probability that a team will play a given formation). This probability function is defined as $p(A_\beta | \Theta_\beta) \rightarrow \mathbb{R} $ which represents the probability that a team $T_\beta$ will select a given action in the set $A_\beta$ (a team formation) and a style type from the set $\Theta_\beta$. 

The payoff function in our game is used to represent the probability of gaining a positive result from a game based on the selected actions, as well as the type and prior beliefs of the opposition. We calculate the probability of a win, draw or loss for a team and weight these to incorporate the positive results. A win is weighted to 2, a draw to 1 and a loss to 0 (so we aim to avoid poor results). The payoff utility function is then defined as $u(a_\alpha, \theta_\alpha | a_\beta, \theta_\beta)) \rightarrow \mathbb{R}$.  This represents the payoff (weighted sum of result probabilities) based on the teams selected actions ($a_\alpha$,$a_\beta$) and their style ($\theta_\alpha$,$\theta_\beta$) where $a \in A$ and the type is $\theta \in \Theta$. We therefore define our Bayesian game as:
\begin{equation}
    G^B = (T, A, \Theta, p, u)\textbf{}
\end{equation}

In this game, we assume that both teams do not know the other teams' tactics. However, both teams have access to the same assumptions using previous data and knowledge on the likely style and formation that a team will use. A team looking to maximise their chances of winning a game would select the action set of decisions which maximises the payoff function and therefore gives the greatest probability of winning a game. However, there are multiple strategies that we can take to optimise the selected decisions depending on the state of the team in real-world (e.g., league position, fighting relegation, a knock-out cup game etc). Therefore, we present three approaches to optimising the selected tactics: 

\begin{itemize}

    \item \textbf{Best Response:} maximises the chances of a win (shown in Equation \ref{eq:best_res}). 
        \begin{equation}\label{eq:best_res}
        \small
            \max \Bigg\{\sum_{a_1 \in A_\alpha} \sum_{a_2 \in A_\beta} u(a_1, \theta_\alpha | a_2, \theta_\beta) \cdot p(a_2|\theta_\beta)\Bigg\}
        \end{equation}
        where, $A_\alpha$ and $A_\beta$ are the set of actions that team $\alpha$ and $\beta$ can take respectively. We aim to maximise the sum of payoffs $u$ multiplied by the probability of the opposition ($T_\beta$) selecting the action $a_2$ and style $\theta_\beta$. This approach has the highest risk as we are not considering the opposition payoff, we just select the best payoff for ourselves. 
        
    \item \textbf{Spiteful Approach:} minimises the chances of losing a game (shown in Equation \ref{eq:min}).
        \begin{equation}\label{eq:min}
        \small
            \min \Bigg\{\sum_{a_1 \in A_\alpha} \sum_{a_2 \in A_\beta} u(a_2, \theta_\beta | a_1, \theta_\alpha) \cdot p(a_2|\theta_\beta)\Bigg\}
        \end{equation}
    where, we aim to minimise the sum of the payoffs $u$ for the opposition team multiplied by the probability of the opposition selecting the action $a_2$ and style $\theta_\beta$. By reducing the chances of the opposition winning the game, this increases the chances of a draw or a win for our team. This approach has the lowest risk as we are not considering our payoff, we are selecting the payoff that limits the opposition.
    
    \item \textbf{Minmax Approach:} In this approach we find the tactics that maximise the chances of winning the game but also minimise the chances of the opposition winning a game (shown in Equation \ref{eq:maxmin}). 
        \begin{equation}\label{eq:maxmin}
        \small
            \max \Bigg\{\sum_{a_1 \in A_\alpha} \sum_{a_2 \in A_\beta} (u(a_1, \theta_\alpha | a_2, \theta_\beta)- u(a_2, \theta_\beta | a_1, \theta_\alpha)) \cdot p(a_2|\theta_\beta)\Bigg\}
        \end{equation}
        
     where, we aim to maximise the sum of the payoffs $u$ for team $\alpha$ while also minimising the sum of the payoffs $u$ for the opposition team. This is then weighted by the probability of the opposition selecting the action $a_2$ and style $\theta_\beta$.
    
\end{itemize}

The different optimisation approaches allow teams to select the tactics which are best suited to their risk levels which may be dependant on the overall position of a team in a league or the individual game scenario. The pre-match decisions that are made by the team are then used as their pre-match tactics which feed into the stochastic game defined next section, where we model the in-match tactical decisions (such as substitutes) so that we can optimise the in-match decisions.  

\subsection{In-Match Stochastic Game}\label{subsec:in-game}

As a game of football progresses the game changes state in-terms of the scoreline, in-turn changing the objectives for either team. If a team is winning they may make defensive changes to ensure they win the game and if a team is losing they may make attacking changes to get back into the game. Due to these state changes, we model the in-game tactical decisions as a stochastic game. 

In our stochastic game, we define the two teams as $T = \{T_\alpha$, $T_\beta\}$ where $T_\alpha$ is the team whose actions we are optimising and $T_\beta$ is the opposing team. We have a set of states $X$ which represent the different possible scorelines in a game starting at 0-0 (where the left number represents the home team goals and the right number represents the away team goal). Each team has a corresponding set of strategies $S_\alpha(x)$ and $S_\beta(x)$ at each of the different states $x \in X$. The strategies represent the current team formation, players and the style of play (the starting strategies are taken from the Bayesian game defined in the previous section). At the starting state $x_0$ (0-0) the team strategies ($S_\alpha(x_0)$ and $S_\beta(x_0)$) correspond to the selected actions from $A_\alpha$ and $A_\beta$ defined in the Bayesian game in the previous section. 

Given the selected strategies of the two teams ($s_1 \in S_\alpha(x)$ and $s_2 \in S_\beta(x)$) and the current state ($x$) we can calculate the probability of a transition to another state $x'$. This is defined as $\pi (x'|x, s_1, s_2)$. In the case of football, from each state there are only two possible states that can be moved to. These will be transitioned by a goal for the home team or a goal for the away team. The other probability we will have to consider is that the state will not be changed for the remainder of the match. In this problem, the length of the game ($t$) is known (90 minutes + injury time) and therefore the probability of state changes will change as the remaining time of the game decreases. The  utility function $u(x, s_1, s_2)$ for this game equates to the probability of a transition into a more positive state (e.g., a team scoring in a 0-0 state to move into a 1-0 state or a winning team (1-0) staying in that state for the remainder of the match time). Given these definitions, we define our stochastic game as:
\begin{equation}
    G^S = (X, T, S(x), \pi, u) 
\end{equation}
Each team aims to move to a more positive state than they are currently in, they make decisions to improve the probability of moving to the more positive state based on their strategy $S_\alpha(x)$. The state changes based on goals in the game, meaning for a game ending 0-0 the state will never move, only time $t$. The example below in Figure \ref{fig:stoch} shows the possible transitions in a game with two goals. We can optimise actions to focus on staying in a positive state (a win) or aiming to move into a more positive state from the current state (e.g., a draw into a win or a loss into a draw).

These stochastic games feed back into future Bayesian games. The future game would have added data to learn from regarding how the decisions made prior performed against certain teams. The transitions made due to choices in the stochastic game will help form beliefs $p(A_\beta|\Theta_\beta)$ regarding what pre-game actions (such as the selected formation) that teams of certain types choose. Also, the two teams in the games will likely play again in the future (teams play each other both home and away each season) and therefore we can learn from our choices and decisions in the first game to improve on in the next game.

\begin{figure}[h!]\centering
\begin{tikzpicture}[>=stealth,shorten >=1pt,auto,node distance=1.75cm]
\node[circle  , draw=black]  (a1) {$0-0$}  ;  
\node[circle  , draw=black, below of= a1, left of = a1](a2)   {$1-0$}  ;
\node[circle  , draw=black,below of= a1, right of = a1](a3)   {$0-1$}  ; 
\node[circle  , draw=black,below of= a2](a4)   {$2-0$ } ; 
\node[circle , draw=black,below of= a3] (a5)  {$0-2$}  ; 
\node[circle , draw=black,left of= a3, below of = a3] (a6)  {$1-1$}  ; 

\path[-latex](a1) edge[loop above] node[sloped]{$\pi_1$}  (a1);
\path[-latex](a1) edge[bend right] node[sloped]{$\pi_2$}  (a2);
\path[-latex](a1) edge[bend left] node[sloped]{$\pi_3$}  (a3);

\path[-latex](a2) edge[loop left] node[sloped]{$\pi_4$}  (a2);
\path[-latex](a2) edge[] node[sloped]{$\pi_5$}  (a4);
\path[-latex](a2) edge[bend left] node[sloped]{$\pi_6$}  (a6);

\path[-latex](a3) edge[loop right] node[sloped]{$\pi_7$}  (a3);
\path[-latex](a3) edge[] node[sloped]{$\pi_8$}  (a5);
\path[-latex](a3) edge[bend right] node[sloped]{$\pi_9$}  (a6);

\path[-latex](a4) edge[loop below] node[sloped]{$\pi_{10}$}  (a4);
\path[-latex](a6) edge[loop below] node[sloped]{$\pi_{11}$}  (a6);
\path[-latex](a5) edge[loop below] node[sloped]{$\pi_{12}$}  (a5);

\begin{pgfonlayer}{background}
    \draw[rounded corners=2.5em,line width=3.5em,red!20,cap=round]
                (a1.center) -- (a2.center) -- (a4.center);
\end{pgfonlayer}

\end{tikzpicture}

\caption{\small An example of a state-transition diagram in a match with 2 goals being scored and the different routes that can be taken through the states. The highlighted route shows the transitions for a match ending 2-0 to the home team.}
\label{fig:stoch}
\end{figure}
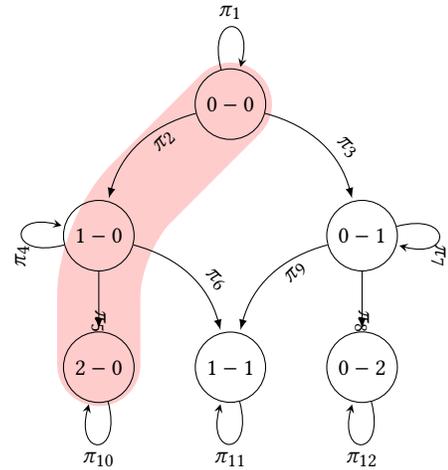

\section{Solving the Pre-Match Bayesian Game}\label{sec:solve-bg}

With the defined game $G^B$ from the previous section, we formulate a model for the pre-match Bayesian game that we solve to select the best tactics which will maximise a team chances of obtaining a positive outcome. To do this there are a number of challenges that we must address. Firstly, we predict the tactics of the opposition by using historical data from similar games so that we can calculate the payoffs of different actions against given opposition actions. Next, we learn the payoff values for the actions that we select and finally we optimise the selected tactics. This is expanded on throughout this section.

\subsection{Predicting the Opposition Strategy}\label{subsec:predict}
When predicting how an opposition will select their strategy, there is limited historical data for games against them in the past. Therefore, we cluster the teams into different playing style categories so we can look for trends in how the opposition play against similar team styles. To cluster the teams we use a feature set containing the number of: passes, shots, goals for, goals against and tackles that a team has made. For the clustering we use a k-means approach for $|C|$ clusters using Equation \ref{eq:cluster} which aims to choose centroids that minimise the inertia, or within-cluster sum-of-squares criterion.
\begin{equation}\label{eq:cluster}
    \sum^{n}_{i=0}\min_{\mu_j \in C}(||x_i-\mu_j||^2)
\end{equation}
where, $n$ is the number of teams and $C$ is the set of cluster means $\mu$.

This allows us to evaluate the style of a team, for example a team with many passes and many shots may be seen as a ``tika-taka'' style team which is an attacking team playing a passing style of football (e.g., the World Cup winning Spain team from 2010 or Barcelona), whereas a team with fewer passes and defensive play may have a ``route one'' style where they look to use long balls over the opposition defence.

Using the clusters of team styles we can learn the strategies that an opposition uses against similar teams. To do this we look at the historical data and build a model using a Support Vector Machine (SVM) with a radial basis function kernel \cite{scholkopf1997comparing} (shown in Equation \ref{eq:rbf}). The algorithm learns using features $x$ which are made up from the tactical set-ups from the prior 5 games against teams from the same style cluster. 
\begin{equation}\label{eq:rbf}
    f(x) = \sum_{i=1}^{C} {\lambda}_{i} \phi (|x - {m}_{i}|) 
\end{equation}
where, $C$ is the clusters, $m$ is the cluster centres and $\lambda$ is the cluster weighting.

\subsection{Learning the Payoffs}\label{subsec:pay-offs}

To learn the payoffs from historical data we develop a model that uses the team's tactical style, potential formation and team strength to give probabilities of a team winning the game. The set of features ($X$) that we use for in our model are: home team style, away team style, home team formation, away team formation and then team strengths are calculated by using the outputs from the model described in \cite{Dixon_Coles} (likelihood of a home win, draw or away win). The target class is the final result of the game: home team win, away team win or a draw.

Using these features, we train a multi-class classification deep neural network. The neural network is trained using stochastic gradient descent using a categorical cross-entropy loss function (shown in Equation \ref{eq:ccelf}) and a soft-max activation function.
\begin{equation}\label{eq:ccelf}
    -\frac{1}{N}\sum^N_{i=1}\log p_{\textit{model}} [y_i \in O_{y_i}]
\end{equation}

where, $N$ is the number of games that we are using to train the model and $p_{\textit{model}} [y_i \in O_{y_i}]$ is the models probability that $y_i$ is in the class $O$. This model takes the given teams, possible playing styles and possible formations to give a probability of winning, drawing or losing the game. Finding and selecting optimised tactics is discussed in the next subsection. 

\subsection{Optimising Pre-Match Tactics}\label{subsec:optimising}

Once we have a model that learns the expected payoffs from the different possible actions (by ourselves and the opposition), we then look to find the best actions/decisions to make, i.e., those which maximise the chances of gaining a positive outcome the game. 

 Firstly, we use the methods that we discuss in Section \ref{subsec:predict} to predict the actions and style that an opposition is likely to select. We use the clustering methods to find their most likely tactical style and then the formation prediction model to give the formation with the highest probability of being selected. By using the predicted opposition style and formation, we explore our possible actions to select the best tactics. Table \ref{tab:bayes_nash} shows the payoffs for the different actions that we can take (when facing a given opposition formation and style). Here, $S$ corresponds to a given style we are able to play in ($x$ possible styles), $f$ corresponds to a given formation ($y$ possible) and then $p(h,d,a|S,f)$ is the probability (output from the model discussed in Section \ref{subsec:pay-offs}) for a home win $h$, draw $d$ and away win $a$ given the selected style and formation. The payoff for the team is the weighted sum of win and draw probabilities and these values are pre-computed so that we can then use the three approaches defined in Section \ref{subsec:pre-game} (best response, spiteful and minmax) to optimise the tactical decisions that we can take depending on the opposition.

\begin{table}[h!]
\begin{tabular}{cccc}
     & $S_1$ & $\hdots$  & $S_x$   \\ \cline{2-4} 
\multicolumn{1}{l|}{$f_1$} & \multicolumn{1}{l|}{$p(h,d,a|S_1,f_1)$} & \multicolumn{1}{l|}{$\hdots$} & \multicolumn{1}{l|}{$p(h,d,a|S_x,f_1)$} \\ \cline{2-4} 
\multicolumn{1}{l|}{$f_2$} & \multicolumn{1}{l|}{$p(h,d,a|S_1,f_2)$} & \multicolumn{1}{l|}{$\hdots$} & \multicolumn{1}{l|}{$p(h,d,a|S_x,f_2)$} \\ \cline{2-4} 
\multicolumn{1}{l|}{$f_3$} & \multicolumn{1}{l|}{$p(h,d,a|S_1,f_3)$} & \multicolumn{1}{l|}{$\hdots$} & \multicolumn{1}{l|}{$p(h,d,a|S_x,f_3)$} \\ \cline{2-4} 
\multicolumn{1}{l|}{$\vdots$} & \multicolumn{1}{l|}{$\vdots$} & \multicolumn{1}{l|}{$\hdots$} & \multicolumn{1}{l|}{$\vdots$} \\ \cline{2-4} 
\multicolumn{1}{l|}{$f_y$} & \multicolumn{1}{l|}{$p(h,d,a|S_1,f_y)$} & \multicolumn{1}{l|}{$\hdots$} & \multicolumn{1}{l|}{$p(h,d,a|S_x,f_y)$} \\ \cline{2-4} 
\end{tabular}
\caption{\small An example payoff table for a team who can have a tactical style of $S_1$ to $S_x$ and a given formation $f_1$ to $f_y$.}
\label{tab:bayes_nash}
\end{table}
\vspace*{-\baselineskip}

\section{Solving the In-Match Stochastic Game}\label{sec:in-game}

We compute optimised strategies for our in-match stochastic game $G^S$ by using historical data of the team tactical setups (style and formation as discussed in the previous section). By using this we learn the state transition probabilities ($\pi$) and evaluate how certain tactical actions will affect this and therefore learn the action payoff values. This would allow teams to make in-match decisions that can boost the chances of staying in a positive state or moving into a more positive state by scoring a goal. An example of this problem is shown in Figure \ref{fig:stoch}. We next detail how we learn the state transition probabilities, action payoffs and optimisation.

\subsection{Learning the State Transition Probabilities}

Prior work by Dixon and Robinson \cite{dixon1998birth} models how the rate of scoring goals changes over the course of a match.  Their model incorporates parameters for both the attacking and the defensive strength of a team, home advantage, the current score and the time left to play. They show how the scoring rate tends to increase over the game but is also influenced by the current score. They then use their model to find the probability that the game will end in a given state which can be used for match outcome prediction and goal-time prediction. We take inspiration from the model presented by Dixon and Robinson to learn the state transition probabilities ($\pi$) that we need to use in our stochastic game.

To learn our state transition probabilities we build a new model at each game-state that will give the probability of each of the possible outcomes from that state (home goal, away goal, no goals). We use a feature set made up from the team strength and the teams' formation and style taken from the Bayesian game (in this game we know our oppositions tactics and style but not the in-match actions they may take). For our model ($\phi$) we use the SVM classification model (with an RBF kernel) described in Section \ref{subsec:predict} and Equation \ref{eq:rbf}. Also, $\pi$ is the transition probability of moving from state $S_x$ to state $S_y$ and $F$ is the feature set. This means $\pi_{x,y} = p(S_x \rightarrow S_y) = \phi_x(F)$ giving a prediction model for each of the possible states $x$ the game could be in.     

\subsection{Learning the Action Payoffs}

We build on the models that we discussed in the previous section to include new features into the feature set $F$ to help us model the effect of in-match decisions such as substitutes. The first new feature we use is a measure of a new player's contribution to the team which represents the impact of a new substitution on the pitch. This allows us to calculate the payoff of the action (substitute) so that we can make an optimised decision at a given point in the game. To calculate the contribution of the players on the bench we use the centrality metric that is discussed in \cite{beal2019teams}. This metric gives the importance of a player to the overall team network.  We also use the remaining match time as a feature so we can see how long an action has to make an impact on the game. These new features are used to update Equation \ref{eq:rbf}. The payoff of each action is the transition probability of moving to a more positive state (e.g., if a team is winning 1-0 it is the probability of making it 2-0 or if a team is losing 3-0 it is the probability of making the game 3-1). 



\subsection{Optimising In-Match Tactics}

Assuming the standard rules of football, each team can make up to 3 substitutions in a game (these can be one at a time or all at once) and has 7 players to choose from, meaning there are 64 combinations of actions (including doing nothing) that we can take at each game-state. We pre-compute the payoffs for each of these possibilities and then select the optimised action to take. Depending on if the team wants to remain in or move to a better state, we can optimise the actions by using two different approaches: 

\begin{itemize}
    \item \textbf{Aggressive approach:} Choose the action that maximises the probability of moving to a more positive state.
    \item \textbf{Reserved approach:} Choose the action that maximises the chances of staying in the current state (if winning).
\end{itemize}

In the next section, we test our learning and optimisation approaches for both the Bayesian and stochastic game discussed in the prior sections. 

\section{Empirical Evaluation}\label{sec:experiments}

To evaluate our models we use a dataset collected from two seasons (2017/18 and 2018/19) from the English Premier League (EPL).\footnote{All data provided by StatsBomb - www.statsbomb.com.} The dataset breaks down each of the games from the tournament into an event-by-event analysis where each event gives different metrics including: event type (e.g., pass, shot, tackle etc.), the pitch coordinates of the event and the event outcome. This type of dataset is industry-leading in football and used by top professional teams.  Thus, this is a rich real-world dataset that allows us to rigorously assess the value of our model. The experiments\footnote{Tests have been run using Scikit-Learn and TensorFlow.} performed are as follows:

\subsection{Experiment 1: Testing the Opposition Action Prediction}

In our first test, we aim to evaluate the performance of the style clustering methods and the team formation prediction. This allows us to accurately predict the tactics of opposition and therefore optimise ours around this.

To select the number of clusters that we use in our testing we use an elbow approach to find the point where the within-cluster sum of squared errors (SSE) will decrease and find that 4 clusters are the best to use. We show in Figure \ref{fig:cluster} how the teams were split into their clusters (based on the 2017/18 season using two of the features for the axis). This gives a view of how the team styles within the league correspond to performance. The triangle shows how Manchester City were far beyond any other team that season and therefore are shown as their own cluster. 

\begin{figure}[h!]
    \centering
    \begin{tikzpicture}
        \begin{axis}[%
            xlabel=Goals For,
    		ylabel=Goals Against,width=\columnwidth-40,
            height=\columnwidth-120,
            y label style={at={(axis description cs:0.15,.5)},anchor=south},
        	scatter/classes={%
        		a={mark=square*,blue},%
        		b={mark=triangle*,red},%
        		c={mark=o,draw=black},
        		d={mark=*, fill=black}}]
        	\addplot[scatter,only marks,%
        		scatter src=explicit symbolic]%
        	table[meta=label] {
                x   y   label
                44 58 a
                28 56 c
                39 47 a
                56 60 c
                44 64 c
                31 56 a
                36 39 a
                35 68 a
                106 27 b
                48 68 a
                74 36 d
                28 58 a
                74 51 d
                45 61 c
                62 38 d
                34 54 a
                37 56 c
                45 55 a
                68 28 d
                84 38 d
        	};
        \end{axis}
    \end{tikzpicture}
    \caption{\small 2017/18 EPL Team Style Clusters.}
    \label{fig:cluster}
\end{figure}
\vspace*{-\baselineskip}

We next test our opposition formation prediction model as discussed in Section \ref{subsec:predict}. Using features taken from the prior 5 games against teams in the same style cluster to ourselves we predict team formation. When testing the model we predict the correct formation with an accuracy of 96.21\% (tested on using a train-test split of 70\% to 30\% with a cross-validation approach for 10 folds). The model achieved a precision score of 0.9867, recall score of 0.9135 and an F1 score of 0.9441. There were a total of 30 different formations used across the season with the most popular formation being `4-2-3-1' used 21\% of the time. 

\subsection{Experiment 2: Learning the Payoffs}

\begin{figure}[h!]
\centering
\begin{tikzpicture}
    \centering
    \begin{axis}[
            ybar,
            bar width=0.3cm,
            symbolic x coords={Accuracy,Precision,Recall,F1 Score},
            xtick=data,
            ylabel={\small Percentage (\%)},
            width=\columnwidth-30,
            height=\columnwidth-120,
            ymin=0,ymax=100,
            y label style={at={(axis description cs:0.15,.5)},anchor=south},
        ]
        \addplot[pattern=north east lines, pattern color=blue, 
            every node near coord/.style={inner ysep=5pt}, 
            error bars/.cd, 
                y dir=both, 
                y explicit] 
        table[x=interval,y=diff]{\mydata};
        \addplot[pattern=horizontal lines, pattern color=red, 
            every node near coord/.style={inner ysep=5pt}, 
            error bars/.cd, 
                y dir=both, 
                y explicit] 
        table[x=interval,y=diff]{\newdata};
    \addlegendentry{\small Payoff Model}
	\addlegendentry{\small Dixon and Coles}
    \end{axis}
\end{tikzpicture}
\caption{\small Payoff Model Performance Comparison.}
\label{fig:payoff}
\end{figure}
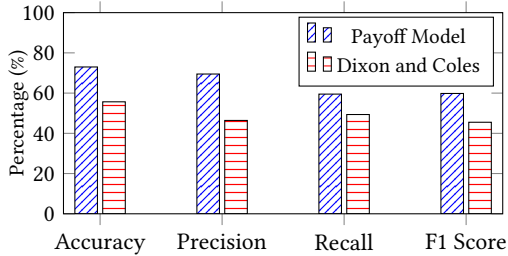

Using the methods discussed in Section \ref{sec:solve-bg}, we build a deep learning model\footnote{We use a fully-connected feed forward NN with 3 layers \& a ReLU activation function.} that takes the actions of the teams and the team strengths into account, the model then assigns a probability (which is used as part of the payoff) to the predicted outcome of the game (home win, draw, away win). We test the outcome probability model by evaluating the accuracy of the model for predicting games in the EPL from the past 2 seasons and comparing our results to those from a well-known football prediction algorithm presented by Dixon and Coles in \cite{Dixon_Coles}. The results from this testing are shown in Figure \ref{fig:payoff} (tested on using a train-test split of 70\% to 30\% with a cross-validation approach for 5 folds).

\pgfplotstableread[row sep=\\,col sep=&]{
    interval & diff  \\
    Accuracy     & 72.99  \\
    Precision     & 69.48  \\
    Recall    & 59.5 \\
    F1 Score   & 59.82  \\
    }\mydata
    
\pgfplotstableread[row sep=\\,col sep=&]{
    interval & diff  \\
    Accuracy     & 55.67  \\
    Precision     & 46.41  \\
    Recall    & 49.35 \\
    F1 Score   &  45.51 \\
    }\newdata

\vspace*{-\baselineskip}
This shows that by having extra information regarding the team formation and style clusters we predict the outcome more accurately and therefore produce better payoffs which are used to optimise our actions in the Bayesian game. 

\subsection{Experiment 3: Pre-Match Optimisation}

To test the pre-match Bayesian game we iterate through all games in our dataset (760 games) across the two EPL seasons and find the optimised tactics using the 3 different optimisation approaches discussed in Section \ref{subsec:pre-game}. 

By calculating the optimised tactics we can compare our approaches and validate our models using real-world data by focusing on a number of metrics. Firstly, we look at how ``close'' our optimised tactics are to what was selected for the real-world game. We define ``closeness'' as a formation that is equal to our recommendation or is only 1 change away (e.g., 4-4-2 is close to 4-5-1 as you can move a striker to midfield to give the ``close'' formation). Using this metric we evaluate the optimisation methods for tactic recommendations and find that the best response method has a closeness of 35.3\%, the spiteful approach has a closeness of 59.7\% and the minmax approach is at 44.6\%. This shows that the spiteful approach is the closest representation to real-world selections. However, when this is split into home and away (50\% and 69\%) tactics we see that this number is skewed by the number of teams that aim to minimise the chances of losing (using the spiteful approach) in away games, this is expanded on in Section \ref{sec:discussion}. 

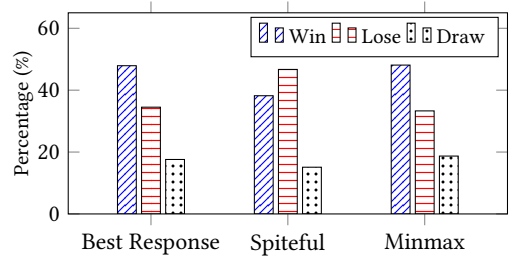
\begin{figure}[h!]
\centering
\begin{tikzpicture}
    \centering
    \begin{axis}[
            ybar,
            legend columns=3, 
            bar width=0.25cm,
            enlarge x limits={abs=1.1cm},
            symbolic x coords={Best Response,Spiteful, Minmax},
            xtick=data,
            ylabel={\small Percentage (\%)},
            width=\columnwidth-30,
            height=\columnwidth-120,
            ymin=0,ymax=65,
            y label style={at={(axis description cs:0.15,.5)},anchor=south},
        ]
        \addplot[pattern=north east lines, pattern color=blue, 
            every node near coord/.style={inner ysep=5pt}, 
            error bars/.cd, 
                y dir=both, 
                y explicit] 
        table[x=interval,y=diff]{\mydata};
        \addplot[pattern=horizontal lines, pattern color=red, 
            every node near coord/.style={inner ysep=5pt}, 
            error bars/.cd, 
                y dir=both, 
                y explicit] 
        table[x=interval,y=diff]{\newnewdata};
        \addplot[pattern=dots, pattern color=black,
            every node near coord/.style={inner ysep=5pt}, 
            error bars/.cd, 
                y dir=both, 
                y explicit] 
        table[x=interval,y=diff]{\newdata};
    \addlegendentry{\small Win}
	\addlegendentry{\small Lose}
	\addlegendentry{\small Draw}
    \end{axis}
\end{tikzpicture}
\caption{\small Percentage of Real-World Results with Close Tactic Selection.}
\label{fig:close}
\end{figure}

We next look at how the team performed in the real-world when the selected tactics were ``close'' to our recommendation. In Figure \ref{fig:close} we show how the results of teams who use our recommendation in terms of the win, draw and loss percentage. This shows that when teams take the minmax approach they are more likely to win a game in comparison to the other approaches (0.2\% more than the best response approach). Although their results are similar, in comparison to the best response, minmax boosts the chance of a draw by 1.1\% and reduces the chance of a loss by 1.2\%. 

\pgfplotstableread[row sep=\\,col sep=&]{
    interval & diff  \\
    Best Response     & 47.9  \\
    Spiteful    & 38.2 \\
    Minmax    & 48.1 \\
    }\mydata
    
\pgfplotstableread[row sep=\\,col sep=&]{
    interval & diff  \\
    Best Response     & 17.6  \\
    Spiteful    & 15.1 \\
    Minmax    & 18.7 \\
    }\newdata
    
\pgfplotstableread[row sep=\\,col sep=&]{
    interval & diff  \\
    Best Response     & 34.5  \\
    Spiteful    & 46.7 \\
    Minmax    & 33.3 \\
    }\newnewdata

\vspace*{-\baselineskip}
Finally, we assess the difference between the payoff of the recommended tactics and the actual tactics used across all 760 games. We find that by taking the best response approach this boosts a teams probability of winning on average by 16.1\% and the minmax approach boosts by 12.7\%, while the spiteful approach reduces the chances of losing a game by 1.4\%. This shows that, as expected, the best response gives the biggest boost to the probability of winning a game, although the minmax approach achieves similar results while also reducing the chances of losing the game. Therefore, the minmax approach would be the best option a team should take in most occasions (depending on the factors of the specific game). 

\subsection{Experiment 4:  Predicting State Transitions}

\begin{figure}[h!]
    \centering
    \includegraphics[scale=0.45]{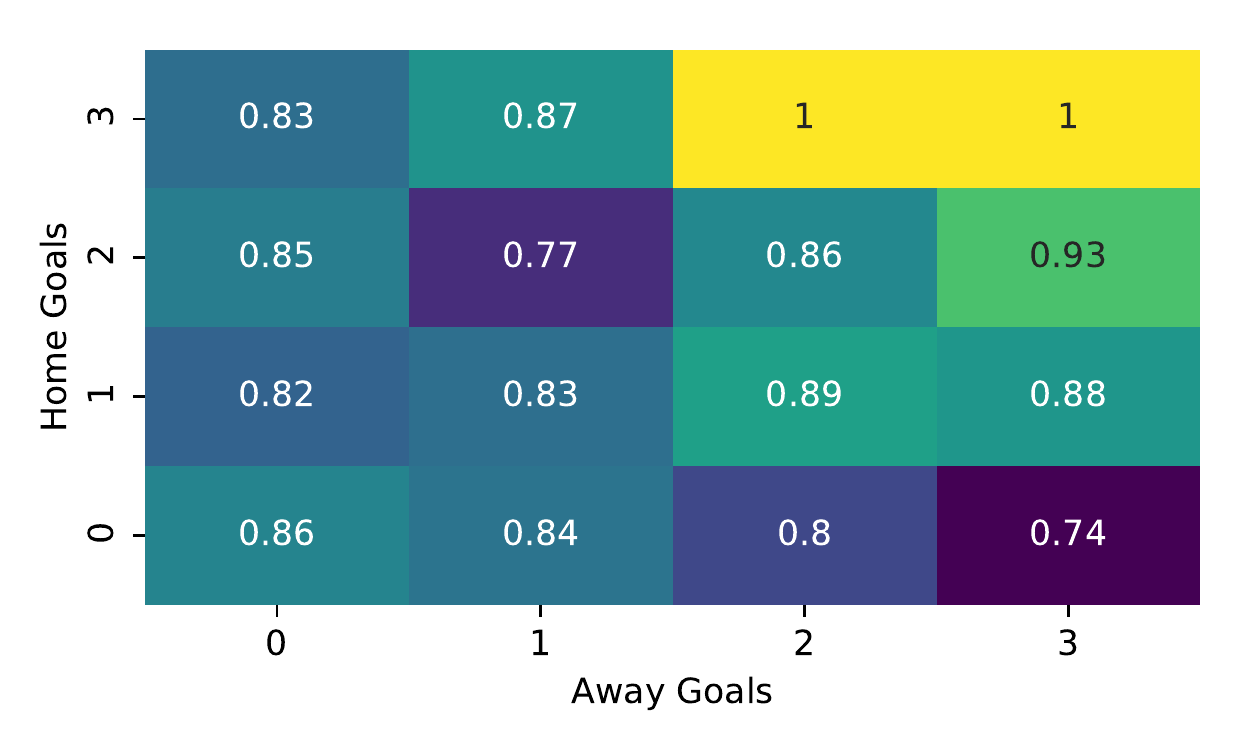}
    \caption{\small Heatmap of State-Transition Model Accuracy.}
    \label{fig:states}
\end{figure}

To test the accuracy of the state transition models (one for each game-state) discussed in Section \ref{sec:in-game}, we compare the model output (home goal, away goal or no goals) to the real-world outcome. We use a train-test split of 70\% to 30\% with a cross-validation approach for 10 folds. We assess each of the models separately using this approach and on average achieve an accuracy of 87.5\% (with a standard deviation of 4.8\%), the detailed results for each of the different states are shown in Figure \ref{fig:states}.\footnote{This has been limited to 3-3 due to the smaller sample sizes of larger scorelines.}

This shows how our models effectively learn the state transition probabilities to a high accuracy at each state. The lower scoreline states have more data points over the last two EPL seasons which we use to train and test the models. Therefore, we have a higher certainty over these state transition models in comparison to the ones trained for the higher scorelines that rarely occur in the real-world (more than 6 goals in a match), hence they are not shown in Figure \ref{fig:states} but are available to use in our next experiment. 

\vspace*{-\baselineskip}

\subsection{Experiment 5: In-Game Optimisation}

When testing the decisions made using the methods from Section \ref{sec:in-game}.3 we iterate through all games in our dataset (760 games) across the two EPL seasons, calculating the payoffs of the actions that both teams can take at each game-state. We compare how often teams took our optimised action in the real world (based on the two different approaches suggested) and if not, evaluate how much our action suggestion would have boosted the team's in-game chances of moving to a more positive state and to win the game.\footnote{We do not have data for the players that are included as substitutes so we consider all squad players (instead of just the 7 substitutes) which impacts our accuracy.}

We first test the action payoff model discussed in Section \ref{sec:in-game}.2 which uses the state transition probability, substitution and the time of the game to calculate the payoff of the given substitute. By so doing, our model (tested in the previous subsection) predicts the next state with an average accuracy is 95.5\% (standard deviation of 4.5\%), again tested using a train-test split of 70\% to 30\% with a cross-validation approach for 10 fold. 

\begin{figure}[h!]
\centering
\begin{tikzpicture}[thick,scale=1, every node/.style={scale=0.75}]
\begin{axis}[ 
xbar, xmin=0.2,xmax=0.5,
xlabel={Average Payoff},
bar width=0.25cm,
symbolic y coords={
    {Aggressive},
    Reserved},
ytick=data,
width=\columnwidth-45,
height=\columnwidth-145,
enlarge y limits={abs=0.5cm},
legend style={at={(0.675,0.05)},anchor=south west}
]
\addplot[pattern=north east lines, pattern color=blue] coordinates {
    (0.3123552084,{Aggressive})
    (0.37482401,Reserved)};
\addplot[pattern=vertical lines, pattern color=red] coordinates {
    (0.32985439,{Aggressive})
    (0.40872,Reserved)};
\addlegendentry{\small Real-World}
\addlegendentry{\small Optimised}
\end{axis}
\end{tikzpicture} 
\caption{\small Payoffs of Real-World vs. Optimised Decisions}
\label{fig:boost}
\end{figure}
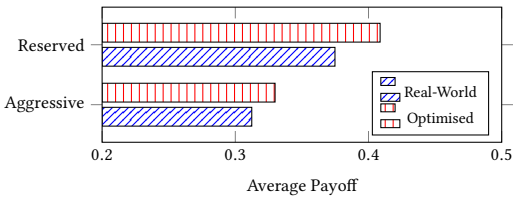

When comparing our action recommendations to those that were taken by the managers in the real-world, we find that the aggressive approach makes the same decision 14.75\% of the time and the reserved approach makes the same decision 14.11\% of the time. If we look at teams making similar player substitutions to our recommendation (selecting a player who plays in the same position as our recommended substitution) then these increase to 40.10\% and 39.75\% respectively. In Figure \ref{fig:boost}, we show the average payoff for each substitute comparing the real world and our two approaches. 

\vspace*{-\baselineskip}

On average our more aggressive approach boosts winning payoffs by 2.0\% and the more reserved approach reduces the opponents winning payoff by 3.4\%. This shows that the changes in tactics that are made in a game can have an impact on the overall outcome and help teams to move into more positive states or stay in the current state if a team is winning a game. By using the stochastic game payoffs we can optimise the efficiency of these decisions by 3.4\% which could have a significant difference to a team across a season in a game such as football, where every marginal gain counts. 



\section{Discussion}\label{sec:discussion}

One key observation  from our testing is how attitude differs between home and away teams. In terms of ``closeness'', the most accurate actions for away teams tactics are given by the spiteful approach; 69\% in comparison to 33\% and 32\% for the best response and minmax respectively. This shows that away teams are more likely to select tactics that minimise the chances of the opposition winning rather than trying to maximise their chances of winning the game. By comparison, the closeness accuracies for home teams are 38\% for best response, 50\% for spiteful and 53\% minmax showing an increase in the number of teams that aim to win a game. This is expected as there is an advantage of playing at home, therefore they chose to minimise their risk of losing.  Overall, teams can increase their chances of winning by up to 16.1\% (p-value of 0.0001).

We also see that when using the spiteful approach (which aims to minimise the risk of losing a game), there is a drop in the chances of a team drawing a match and it increases the chances of losing the match in comparison to the other approaches (shown in Figure \ref{fig:close}). The key reason for this is likely due to that when teams aim to avoid losing and pick a more defensive approach, this allows the opposition to attack more and have more chances to win the game. It also suggests that if a team selects this approach they do not believe they can win a game, as they may be playing an opposition with better players and a higher chance of winning regardless of the selected tactic.

Testing the stochastic game is more challenging due to a lack of data for available substitutes. There is also greater uncertainty regarding the state transitions probabilities. However, we show that by using our modelling approach teams could increase their chances of moving into (or staying in) a more positive state by up to 3.4\% (p-value of 0.0004). 

\section{Conclusions}\label{sec:conclusions}

This paper presents a novel model for making both pre-match and in-match decisions for football; and as such provides a natural framework to conduct learning in. For the pre-match Bayesian game setting, we find that we can effectively predict the style and actions of the opposition and then present three models for selecting optimised actions in response. We find that a minmax approach would be the best to take, however in the real-world teams tend to go for a spiteful approach. Overall, the Bayesian game model could be useful to help real-world teams make effective decisions to win a game and the stochastic game can help coaches/managers make optimised changes during the 90 minutes of a match. In future work, we would break down football and its state to learn more about how the game develops. We could explore similar approaches to the reinforcement learning methods used in AlphaGo \cite{silver2016mastering} to gain a deeper understanding of the in-game dynamics. 

\section*{Acknowledgements}

We would like to thank the reviewers for their comments. This research is supported by the AXA Research Fund and the EPSRC NPIF doctoral training grant number EP/S515590/1.


\bibliographystyle{ACM-Reference-Format}  
\bibliography{sample-bibliography}  

\end{document}